\documentclass[conference]{IEEEtran}
\usepackage[LGR,T1]{fontenc}
\usepackage[latin9]{inputenc}
\usepackage{array}
\usepackage{rotating}
\usepackage{booktabs}
\usepackage{multirow}
\usepackage{amsmath}
\usepackage{amssymb}
\usepackage{graphicx}

\makeatletter

\providecommand{\tabularnewline}{\\}

\usepackage{array}
\usepackage{units}
\usepackage{multirow}
\usepackage{xcolor}

\providecommand{\tabularnewline}{\\}

\makeatother

\begin{document}
\title{On the Energy and Communication Efficiency Tradeoffs in Federated
and Multi-Task Learning}
\author{
}
\author{\IEEEauthorblockN{Stefano Savazzi, Vittorio Rampa, Sanaz Kianoush}
\IEEEauthorblockA{Consiglio Nazionale delle Ricerche (CNR)\\
 IEIIT institute, Milano\\
 Email: \{name.surname\}@ieiit.cnr.it} \and \IEEEauthorblockN{Mehdi Bennis} \IEEEauthorblockA{Centre for Wireless Communications\\
 University of Oulu, Finland\\
 Email: mehdi.bennis@oulu.fi}}
\maketitle
\begin{abstract}
Recent advances in Federated Learning (FL) have paved the way towards
the design of novel strategies for solving multiple learning tasks
simultaneously, by leveraging cooperation among networked devices.
Multi-Task Learning (MTL) exploits relevant commonalities across tasks
to improve efficiency compared with traditional transfer learning
approaches. By learning multiple tasks jointly, significant reduction
in terms of energy footprints can be obtained. This article provides
a first look into the energy costs of MTL processes driven by the
Model-Agnostic Meta-Learning (MAML) paradigm and implemented in distributed
wireless networks. The paper targets a clustered multi-task network
setup where autonomous agents learn different but related tasks. The
MTL process is carried out in two stages: the optimization of a \emph{meta-model}
that can be quickly adapted to learn new tasks, and a\emph{ task-specific
model adaptation} stage where the learned meta-model is transferred
to agents and tailored for a specific task. This work analyzes the
main factors that influence the MTL energy balance by considering
a multi-task Reinforcement Learning (RL) setup in a robotized environment.
Results show that the MAML method can reduce the energy bill by at
least $2\times$ compared with traditional approaches without inductive
transfer. Moreover, it is shown that the optimal energy balance in
wireless networks depends on uplink/downlink and sidelink communication
efficiencies.
\end{abstract}

\IEEEpeerreviewmaketitle{}

\section{Introduction}

The underlying premise of Federated Learning (FL) is to train a distributed
and privacy-preserving Machine Learning (ML) model for resource-constrained
devices \cite{drl-1,key-3,key-4,jointopt}. Typically, FL requires
frequent and intensive use of communication resources to exchange
model parameters with the parameter server \cite{key-5}. However,
it is not optimized for incremental model (re)training and tracking
changes in data distributions, or for learning new tasks, i.e., Multi-Task
Learning (MTL). In particular, jointly learning new tasks using prior
experience, and quickly adapting as more training data becomes available
is a challenging problem in distributed ML and mission critical applications
\cite{commmag}, a topic that still remains in its infancy \cite{AutoML}. 

\begin{figure}[!t]
\centering \includegraphics[scale=0.36]{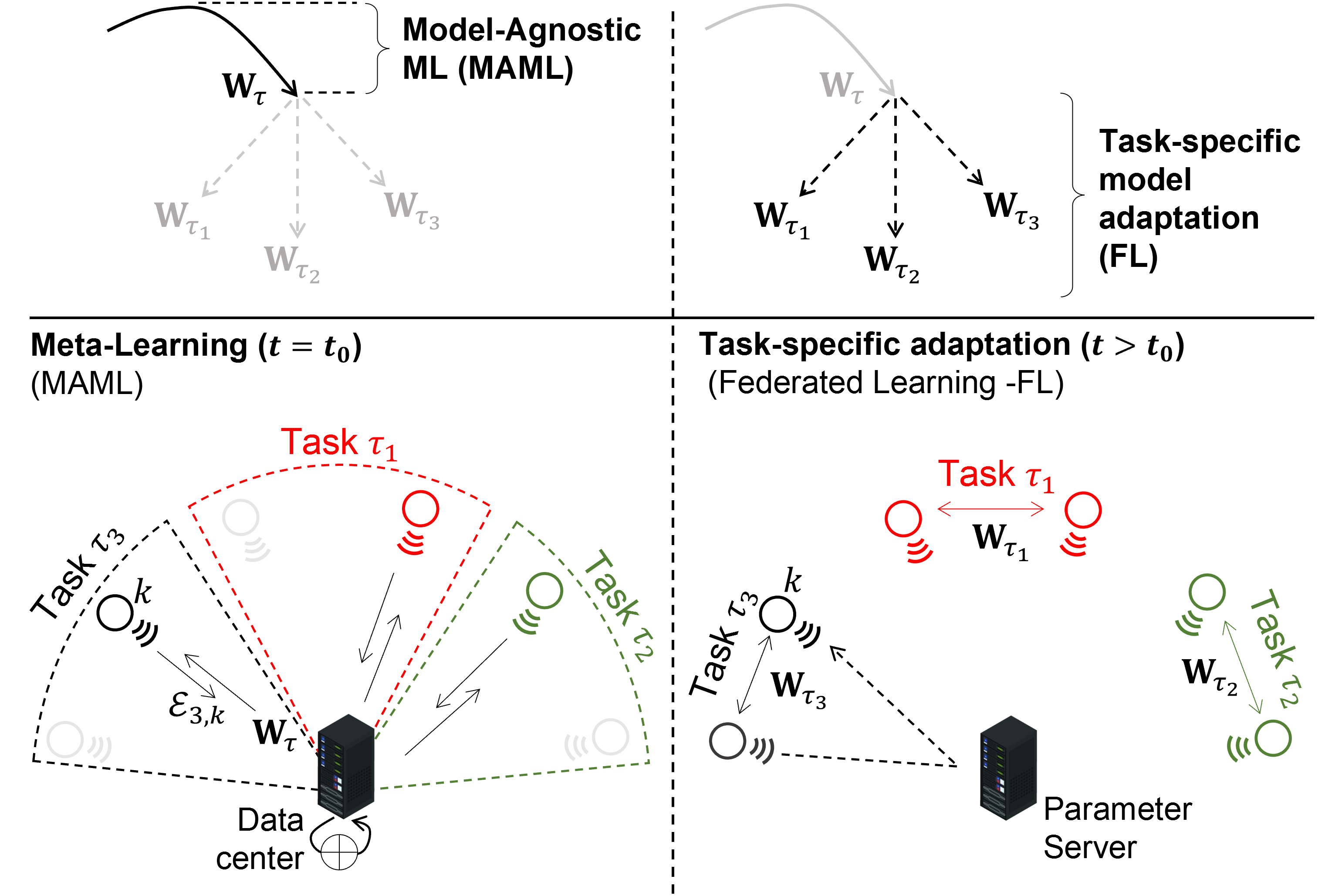} 
 \protect\caption{\label{intro} Clustered multi-task wireless network example. From
left to right: Model-agnostic Meta-Learning (MAML) on the data center,
and decentralized Federated Learning (FL) for task-specific model
adaptation .}
\vspace{-0.4cm}
 
\end{figure}
To obviate this problem, meta-learning is a promising enabler in multi-task
settings as it exploits commonalities across tasks. In addition, meta-learning
relies on the optimization of a ML \emph{meta-model }that can be quickly
adapted to learn new tasks from a small training dataset by utilizing
few communication and computing resources (few shot learning) \cite{maml}.
As depicted in Fig. \ref{intro}, the meta-learning process requires
an initial training stage ($t_{0}$ rounds) where a meta-model $\mathbf{W}_{\tau}$
is trained on a data center, using observations from different tasks,
and fed back to devices (inductive transfer) for task-specific adaptations.
The meta-model $\mathbf{W}_{\tau}$ could be quickly re-trained on
the devices in a second stage ($t>t_{0}$), to learn a task-specific
model $\mathbf{W}_{\tau_{i}}$, optimized to solve the (new) task
$\tau_{i}$. Task-specific training of $\mathbf{W}_{\tau_{i}}$ can
be implemented via FL using few communication resources and a small
amount of data from the new task(s) \cite{IML}. Considering the popular
Model-Agnostic Meta-Learning (MAML) algorithm \cite{maml}, the meta-model
$\mathbf{W}_{\tau}$ is optimized inside a data center using training
data from a subset of devices (\emph{e.g.,} data producers such as
sensors, machines and personal devices) and tasks. The meta-optimizer
is typically gradient-based, and alternates task-specific model adaptations
and meta-optimization stages on different data batches \cite{maml_osvaldo}.
Although the optimization of the meta-model could be energy-hungry
requiring more learning rounds than conventional ML on single tasks,
the meta-learning process incurs lower energy consumption of subsequent
task-specific model updates. Characterizing the tension between meta-model
optimization and task-specific adaptation is currently overlooked
and constitutes the focus of this work.

\textbf{Contributions:} this work provides a first look into the energy
and communication footprints of meta-learning techniques implemented
in distributed multi-task wireless networks. In particular, we consider
a framework that quantifies the end-to-end energy cost of MAML optimization
inside a data center, as well as the cost of subsequent task-specific
model adaptations, implemented using FL \cite{first_look,key-7}.
The work also includes, for the first time, comparisons and trade-off
considerations about MAML-based optimization, and conventional transfer
learning tools over resource-constrained devices. The framework is
validated by targeting a multi-task Reinforcement Learning (RL) setup
using real world data. 

The paper is organized as follows. Sects. \ref{sec:Energy-footprint-modeling}
and \ref{sec:Decentralized-FL:-gossip} describe the meta-learning
approach, as well as its energy footprint evaluation, with a particular
focus on communication and computing costs. In Sect. \ref{sec:A-case-study},
we consider a case study in a real-world industrial workplace consisting
of small networked robots collaboratively learning an optimized sequence
of motions, to follow an assigned task/trajectory. Finally, Sect.
\ref{sec:Conclusions} draws some conclusions.

\section{Multi-task learning and network model\label{sec:Energy-footprint-modeling}}

The learning system under consideration consists of a clustered multi-task
network of $K$ devices and one data center ($k=0$) co-located with
a core network access point i.e, gateway, or Base Station (BS). The
wireless network leverages UpLink (UL) and DownLink (DL) communication
links with the data center (BS), or direct, i.e. sidelink (SL), communications.
As depicted in Fig. \ref{intro}, the devices, or agents, are allowed
to cooperate with each other to learn $M\ll K$ distinct tasks $\tau_{1},...,\tau_{M}$
and form $M$ clusters $\mathcal{C}_{i}$ ($i=1,...,M$). The objective
of the agents in the cluster $\mathcal{C}_{i}$ ($k\in\mathcal{C}_{i}$)
is to learn a task $\tau_{i}$ by estimating the model parameters
$\mathbf{W}_{\tau_{i}}$ through the minimization of a finite-sum
objective function $\mathcal{L}_{i}(\mathbf{W})$ of the general form

\begin{equation}
\mathbf{W}_{\tau_{i}}=\underset{\mathbf{W}}{\mathrm{argmin}}\thinspace\mathcal{L}_{i}(\mathbf{W})=\underset{\mathbf{W}}{\mathrm{argmin}}\underset{\mathcal{L}_{i}(\mathbf{W})}{\underbrace{\sum_{k\in\mathcal{C}_{i}}L_{k}(\mathbf{W}|\mathcal{E}_{i,k})},}\label{eq:fdd}
\end{equation}
where $L_{k}(\mathbf{W}|\mathcal{E}_{i,k})$ is the loss function,
or cost, associated with the $k$-th device $L_{k}(\mathbf{W}|\mathcal{E}_{i,k})=\sum_{\mathbf{x}_{h,i}\in\mathcal{E}_{i,k}}\ell(\mathbf{W}|\mathbf{x}_{h,i})$,
while $\ell(\mathbf{W}|\mathbf{x}_{h,i})$ is the loss function of
the predicted model with data/examples $\mathbf{x}_{h,i}\in\mathcal{E}_{i,k}$
drawn from the task $\tau_{i}$. Notice that the costs of the individual
devices belonging to the same cluster are minimized at the same location
$\mathbf{W}_{\tau_{i}}$ as solving the same task $\tau_{i}$. However,
considering two tasks $\tau_{i},\tau_{j}$ with $i\neq j$, it is
$\mathbf{W}_{\tau_{i}}$ $\neq\mathbf{W}_{\tau_{j}}$. As clarified
in the next section, the devices operate in a streaming data setting
targeting a RL problem: each agent $k$ observes the environment at
each time instant $t$ and obtains samples about the state, actions
and task-dependent rewards. 

The minimization of (\ref{eq:fdd}) is implemented incrementally via
gradient optimization. In particular, we adopt an inductive transfer
learning approach \cite{multitask}. First, we learn a meta-model
$\mathbf{W}_{\tau}$ on the data center using examples drawn from
a subset $\mathcal{Q_{\tau}}$ of $Q\leq M$ tasks $\tau_{i}\in\mathcal{Q_{\tau}}$.
After $t_{0}$ MAML rounds \cite{maml}, the meta-model $\mathbf{W}_{\tau}$
is transferred from the data center to all the individual devices
and adapted to the specific task via FL \cite{commmag}. Notice that
FL is implemented separately by devices in each cluster: in other
words, only the agents within the same cluster are allowed to share
their local models to learn a task-specific global model $\mathbf{W}_{\tau_{i}}$.
Both the optimization of the meta-model and the subsequent task-specific
adaptation stages contribute to the energy cost that are addressed
in Sect. \ref{sec:Decentralized-FL:-gossip}.

\subsection{Model-Agnostic Meta-Learning (MAML)}

Meta-learning operates at a higher level of abstraction \cite{maml_osvaldo}
compared with conventional supervised learning. It uses randomly sampled
observations from different tasks to identify a single model $\mathbf{W}_{\tau}$
such that, once deployed, few training steps are needed to adapt the
meta-model to the new task(s) of interest. As seen in Fig. \ref{intro},
MAML is implemented at the server upon the collection of random data
$\mathcal{E}_{i,k}$ from $Q$ \emph{training tasks} $\tau_{i}\in\mathcal{Q_{\tau}}$
over the uplink. The meta-model is optimized as 
\begin{equation}
\mathbf{W}_{\tau}=\underset{\mathbf{W}}{\mathrm{argmin}}\thinspace\sum_{\tau_{i}\in\mathcal{Q_{\tau}}}\mathcal{L}_{i}(\mathbf{W}),\label{eq:optimizmeta}
\end{equation}
and solved iteratively via gradient-based optimization over multiple
meta-learning rounds (i.e., MAML rounds). 

Each MAML round is divided into two phases, namely the \emph{task-specific
training} stage, followed by the \emph{meta-model update} stage. During
task-specific training stage, $Q$ model adaptations $\mathbf{\mathbf{\varphi}}_{t,\tau_{i}}$
are obtained for each training task $\tau_{i}\in\mathcal{Q_{\tau}}$
using a sub-set $\mathcal{E}_{i,k}^{(a)}$ of the training data $\mathcal{E}_{i,k}^{(a)}\subset\mathcal{E}_{i,k}$
and the Stochastic Gradient Descent (SGD) algorithm. For iteration
$t>0$ with random initialization at $t=0$ and training task $\tau_{i}$
it is
\begin{equation}
\boldsymbol{\mathbf{\varphi}}_{t,\tau_{i}}=\mathbf{W}_{t,\tau}-\mu\times\sum_{k\in\mathcal{C}_{i}}\nabla_{\mathbf{W}_{t,\tau}}L_{k}(\mathbf{W}_{t,\tau}|\mathcal{E}_{i,k}^{(a)}),\label{nofederation}
\end{equation}
where $\mu$ is the SGD step size while $\nabla_{\mathbf{W}_{t,\tau}}L_{k}$
is the gradient of the loss function in (\ref{eq:fdd}) \emph{w.r.t.}
the meta-model $\mathbf{W}_{t,\tau}$. The subsequent meta-model update
stage obtains an improved meta-model $\mathbf{W}_{t+1,\tau}$ for
the next iteration $t+1$ that is trained over the remaining samples
$\mathcal{E}_{i,k}^{(b)}:=\mathcal{E}_{i,k}\setminus\mathcal{E}_{i,k}^{(a)}$
(validation samples),
\begin{equation}
\mathbf{W}_{t+1,\tau}=\mathbf{W}_{t,\tau}-\eta\times\sum_{i=1}^{Q}\sum_{k\in\mathcal{C}_{i}}\nabla_{\mathbf{W}_{t,\tau}}L_{k}\left[\boldsymbol{\mathbf{\varphi}}_{t,\tau_{i}}|\mathcal{E}_{i,k}^{(b)}\right].\label{eq:met}
\end{equation}
Notice that the vector $\nabla_{\mathbf{W}_{t,\tau}}L_{k}\left[\boldsymbol{\mathbf{\varphi}}_{t,\tau_{i}}|\mathcal{E}_{i,k}^{(b)}\right]$
requires gradient-through-gradient operation thus increasing the computational
costs \cite{IML}. In particular, it is
\begin{equation}
\nabla_{\mathbf{W}_{t,\tau}}L_{k}\left[\boldsymbol{\mathbf{\varphi}}_{t,\tau_{i}}|\mathcal{E}_{i,k}^{(b)}\right]=\mathbf{J}_{\mathbf{W}_{t,\tau}}\left[\boldsymbol{\mathbf{\varphi}}_{t,\tau_{i}}\right]\times\nabla_{\mathbf{\varphi}_{t,\tau_{i}}}L_{k}\left[\boldsymbol{\mathbf{\varphi}}_{t,\tau_{i}}|\mathcal{E}_{i,k}^{(b)}\right]\label{eq:grad-grad}
\end{equation}
where $\mathbf{J}_{\mathbf{W}_{t,\tau}}\left[\boldsymbol{\mathbf{\varphi}}_{t,\tau_{i}}\right]$
is the Jacobian operator while $\boldsymbol{\mathbf{\varphi}}_{t,\tau_{i}}=\boldsymbol{\mathbf{\varphi}}_{t,\tau_{i}}(\mathbf{W}_{t,\tau})$
is defined in (\ref{nofederation}). First-order approximation of
$\mathbf{J}_{\mathbf{W}_{t,\tau}}\left[\boldsymbol{\mathbf{\varphi}}_{t,\tau_{i}}\right]\approx\mathbf{I}$
is often used \cite{maml}; in this case, (\ref{eq:grad-grad}) simplifies
as $\nabla_{\mathbf{W}_{t,\tau}}L_{k}\left[\boldsymbol{\mathbf{\varphi}}_{t,\tau_{i}}|\mathcal{E}_{i,k}^{(b)}\right]\approx\nabla_{\boldsymbol{\mathbf{\varphi}}_{t,\tau_{i}}}L_{k}\left[\boldsymbol{\mathbf{\varphi}}_{t,\tau_{i}}|\mathcal{E}_{i,k}^{(b)}\right]$. 

In what follows, the meta-optimization (\ref{eq:met}) is implemented
for $t_{0}$ MAML rounds $\mathbf{W}_{\tau}\cong\mathbf{W}_{t_{0},\tau}$.
As analyzed in the following, the energy footprint of meta-optimization
is primarily ruled by the number of rounds $t_{0}$ as well as the
number of training tasks $Q$ chosen by the meta-optimizer. 

\subsection{Task-specific adaptation via FL network}

Model adaptation uses a decentralized FL implementation \cite{commmag}.
Each device $k$ hosts a model initialized at time $t=0$ using the
meta-model $\mathbf{W}_{t_{0},\tau}$, therefore $\mathbf{W}_{0,\tau_{i}}^{(k)}=\mathbf{W}_{t_{0},\tau}$.
The local model $\mathbf{W}_{t,\tau_{i}}^{(k)}$ is then updated on
consecutive rounds $t>0$ using new data/examples $\mathcal{E}_{i,k}$
drawn from possible new tasks $\tau_{i}$ and then shared with neighbor
devices to implement average consensus \cite{camad}. Defining $\mathcal{N}_{k,i}\subseteq\mathcal{C}_{i}$
as the set that contains $N$ neighbors of node $k\in\mathcal{C}_{i}$,
at every new round ($t>0$) each \emph{k}-th device updates the local
model as 
\begin{equation}
\mathbf{W}_{t+1,\tau_{i}}^{(k)}\leftarrow\mathbf{W}_{t,\tau_{i}}^{(k)}+\sum_{h\in\mathcal{N}_{k,i}}\sigma_{k,h}\cdot(\mathbf{W}_{t,\tau_{i}}^{(h)}-\mathbf{W}_{t,\tau_{i}}^{(k)}),\label{eq:averaging}
\end{equation}
where the weights $\sigma_{k,h}=\left|\mathcal{E}_{i,h}\right|/\sum_{j\in\mathcal{N}_{k,i}}\left|\mathcal{E}_{i,j}\right|$,
computed using the size of the data distributions $\left|\mathcal{E}_{i,h}\right|$
and $\left|\left\{ \mathcal{E}_{i,j}\right\} _{j\in\mathcal{N}_{k,i}}\right|$,
are defined as in \cite{key-5}. The local model update and the consensus
steps are repeated until $\mathbf{W}_{\tau_{i}}=\underset{t\rightarrow\infty}{\lim}\mathbf{W}_{t,\tau_{i}}^{(k)}$,
$\forall k\in\mathcal{C}_{i}$, or a desired training accuracy is
obtained.

\subsection{Reinforcement learning problem}

In the following, we consider, as an experimental case study, a Deep
Reinforcement Learning (DRL) problem and off-policy training\footnote{A random $\varepsilon$-greedy policy is used for gathering experience
in the environment which is independent from the policy $\pi_{i}$
being learned to solve the selected task. Other setups are also possible. }: for task $\tau_{i}$, the training dataset $\mathcal{E}_{i,k}:=(\mathbf{x}_{h,i},y_{h,i},r_{h,i},\mathbf{x}_{h+1,i},y_{h+1,i},r_{h+1,i},...)$
contains state ($\mathbf{x}_{h,i}$), action ($y_{h,i}$), and reward
($r_{h,i}$) time-domain sequences, that generally depend on the chosen
problem $\tau_{i}$ to be solved. A specific example\footnote{In the considered case study, only the rewards values $r_{h,i}$ are
task-dependent, observations while actions follow the same $\varepsilon$-greedy
policy.} is given in Sect. \ref{sec:A-case-study}. In RL jargon, for each
task $\tau_{i}$, the problem we tackle is to learn a policy $\widehat{y}_{h+1,i}=\pi_{i}(\mathbf{x}_{h,i})$
that chooses the best action $\widehat{y}_{h+1,i}$ to be taken at
time $h+1$ given the observation $\mathbf{x}_{h,i}$ at time $h$.
We implemented a Deep Q-Learning (DQL) method: therefore, the policy
is made greedy with respect to an optimal action-value function \cite{deepmind}
$q_{\pi_{i}}^{*}(\mathbf{x}_{h,i},y_{h,i})$, such that $\pi_{i}(\mathbf{x}_{h,i})\doteq\arg\max_{y}q_{\pi_{i}}^{*}(\mathbf{x}_{h,i},y)$.
The function $q_{\pi_{i}}^{*}$ (Q-function) predicts the expected
future rewards for each possible action using the observations $\mathbf{x}_{h,i}$
as inputs. The Q-function $q_{\pi_{i}}^{*}$ is parameterized by a
Deep Neural Network (DNN) model $q_{\pi_{i}}(\mathbf{x}_{h,i},y_{h,i}|\mathbf{W}_{\tau_{i}})$
and learned to minimize the loss function
\begin{equation}
\ell(\mathbf{x}_{h,i}|\mathbf{W}_{\tau_{i}})=\left[r_{h,i}+\nu\max_{y}\left(\widetilde{q}_{\pi_{i}}\right)-q_{\pi_{i}}(\mathbf{x}_{h,i},y|\mathbf{W}_{\tau_{i}})\right]^{2},\label{eq:loss}
\end{equation}
according to the Bellman equation, with $\nu=0.99$ being the discount
factor and $\widetilde{q}_{\pi_{i}}$ a target Q-function network
according to the double learning implementation \cite{ddql}. 

The meta-learning process obtains a meta-model $\mathbf{W}_{\tau}$,
solution to (\ref{eq:optimizmeta}), that best represents the optimal
Q-function set $\left\{ q_{\pi_{i}}^{*},\forall i\text{ s.t. }\tau_{i}\in\mathcal{Q_{\tau}}\right\} $.
Next, the parameters $\mathbf{W}_{\tau}$ could be quickly adapted
during the task-specific adaptation ($t>t_{0}$) to approximate a
specific Q-function $q_{\pi_{i}}^{*}$ solving a (new or unvisited)
task $\tau_{i}$, i.e.,$\tau_{i}\notin\mathcal{Q_{\tau}}$. Notice
that, during the meta-optimization stages, the training data $\mathcal{E}_{i,k}$
for $Q$ selected tasks are moved to the data center at each round
using UL communication. Instead, during task-specific adaptation,
devices from the same cluster $\mathcal{C}_{i}$ implement decentralized
FL and exchange the parameters $\mathbf{W}_{\tau_{i}}$ of their estimated
Q-function, rather than the training sequences: the data-center is
thus not involved since devices communicate via sidelinks. In what
follows, the average running reward $R=\sum_{h}\nu^{h}r_{h,i}$ is
chosen as the accuracy indicator for each task$.$

\section{Energy and communication footprint model}

\label{sec:Decentralized-FL:-gossip}

The total amount of energy consumed by the MTL process is broken down
into computing and communication \cite{key-5}. Both the data center
($k=0$) and the devices ($k>0$) contribute to the energy costs,
although data center is used for meta-model optimization and inductive
transfer only. The energy cost is modelled as a function of the energy
$E_{k}^{(\mathrm{C})}$ required for SGD computation, and the energy
$E_{k,h}^{(\mathrm{T})}$ per correctly received/transmitted bit over
the wireless link ($k,h$). The cost also includes the power dissipated
in the RF front-end, in the conversion, baseband processing and other
transceiver stages \cite{key-5}. In what follows, we compute the
energy cost required for the optimization of the meta-model (ML) and
the task-specific model adaptation (FL), considering $M$ tasks. Numerical
examples are given in Sect. \ref{sec:A-case-study}.

\subsection{Meta-learning and task-specific adaptation}

Training of the meta-model $\mathbf{W}_{\tau}$ runs for $t_{0}$
rounds inside the data center $k=0$. For each round, the data center:\emph{
i)} collects new examples $\mathcal{E}_{i,k}$ from $Q$ training
tasks; \emph{ii)} obtains $Q$ model adaptations (\ref{nofederation})
using data batches $\mathcal{E}_{i,k}^{(a)}\subset\mathcal{E}_{i,k}$,
and \emph{iii)} updates the meta-model for a new round by computing
$Q$ gradients (\ref{eq:met}) using batches $\mathcal{E}_{i,k}^{(b)}\subset\mathcal{E}_{i,k}$.
In particular, the cost of a single gradient computation $E_{0}^{(\mathrm{C})}=P_{0}\cdot T_{0}$
on the data center depends on the GPU/CPU power consumption $P_{0}$
and the time span $T_{0}$ required for processing an individual batch
of data. Defining $\mathrm{B}_{i}^{(a)}$ and $\mathrm{B}_{i}^{(b)}$
as the number of data batches from the training sets $\mathcal{E}_{i,k}^{(a)}$
and $\mathcal{E}_{i,k}^{(b)}$, respectively, the total, end-to-end,
energy (in Joule {[}J{]}) spent by the MAML process is broken down
into learning ($\mathrm{L}$) and communication ($\mathrm{C}$) costs
\begin{equation}
E_{\mathrm{ML}}(t_{0},Q)=E_{\mathrm{ML}}^{(\mathrm{L})}(t_{0},Q)+E_{\mathrm{ML}}^{(\mathrm{C})}(Q).\label{eq:eml}
\end{equation}
 For $t_{0}$ rounds and $Q$ task examples, it is

\begin{equation}
\begin{array}{c}
E_{\mathrm{ML}}^{(\mathrm{L})}(t_{0},Q)=\gamma\cdot t_{0}\cdot\sum_{i=1}^{Q}\sum_{k\in\mathcal{C}_{i}}\left[\mathrm{B}_{i}^{(a)}+\beta\mathrm{B}_{i}^{(b)}\right]E_{0}^{(\mathrm{C})}\\
E_{\mathrm{ML}}^{(\mathrm{C})}(Q)\hspace{-0.1cm}=t_{0}\sum_{i=1}^{Q}\sum_{k\in\mathcal{C}_{i}}b(\mathcal{E}_{i,k})E_{k,0}^{(\mathrm{T})}+\sum_{k=1}^{K}b(\mathbf{W})E_{0,k}^{(\mathrm{T})}
\end{array}\label{eq:cfa-1-1}
\end{equation}
where the sum $\sum_{k=1}^{Q}\sum_{k\in\mathcal{C}_{i}}\mathrm{B}_{i}^{(a)}E_{0}^{(\mathrm{C})}$
quantifies the energy bill for $Q$ task-specific adaptations in (\ref{nofederation}), $\beta\cdot\sum_{i=1}^{Q}\sum_{k\in\mathcal{C}_{i}}\mathrm{B}_{i}^{(b)}E_{0}^{(\mathrm{C})}$
accounts for the meta-model update, and $\beta\geq1$ includes the
cost of the Jacobian computation ($\beta=1$ is assumed under first-order
approximation). $\gamma$ is the Power Usage Effectiveness (PUE) of
the considered data center \cite{cooling}. UL communication of training
data $\mathcal{E}_{i,k}$ has cost $\sum_{i=1}^{Q}\sum_{k\in\mathcal{C}_{i}}b(\mathcal{E}_{i,k})E_{k,0}^{(\mathrm{T})}$
that scales with the data size $b(\mathcal{E}_{i,k})$. DL communication
$\sum_{k=1}^{K}b(\mathbf{W})E_{0,k}^{(\mathrm{T})}$ is required to
propagate the meta-model $\mathbf{W}_{t_{0},\tau}$ to all devices:
$b(\mathbf{W})$ quantifies the size (in Byte) of the meta-model trainable
layers.

Task-specific model adaptations ($t>t_{0}$) are implemented independently
by the devices in each cluster $\mathcal{C}_{i}$ using the same meta-model
$\mathbf{W}_{t_{0},\tau}$ as initialization and the FL network. The
energy footprint for adaptation over the task $\tau_{i}$ ($i=1,...,M$)
could be similarly broken down into learning and communication costs
\cite{key-5} as
\begin{equation}
E_{\mathrm{FL}}(t_{i})=E_{\mathrm{FL}}^{(\mathrm{L})}(t_{i})+E_{\mathrm{FL}}^{(\mathrm{C})}(t_{i})\label{eq:efl}
\end{equation}
with 
\begin{equation}
\begin{array}{c}
E_{\mathrm{FL}}^{(\mathrm{L})}(t_{i})=t_{i}\cdot\sum_{k\in\mathcal{C}_{i}}\mathrm{B}_{i}E_{k}^{(\mathrm{C})}\\
E_{\mathrm{FL}}^{(\mathrm{C})}(t_{i})=b(\mathbf{W})\left[t_{i}\cdot\sum_{k\in\mathcal{C}_{i}}\sum_{h\in\mathcal{N}_{k,i}}E_{k,h}^{(\mathrm{T})}\right].
\end{array}\begin{aligned}\end{aligned}
\label{eq:fa-1}
\end{equation}
$t_{i}$ is the required number of FL rounds to achieve the assigned
average running reward $R$ for the corresponding task, and $\mathrm{B}_{i}$
is the number of data batches from the training set $\mathcal{E}_{i,k}$
(now collected for task-specific model adaptation). Notice that $E_{k,h}^{(\mathrm{T})}$
represent the energy spent for sidelink communication between device
$k$ and device $h\in\mathcal{N}_{k,i}$ in the corresponding neighborhood.
When sidelink communication is not available, direct communication
can be replaced by UL and DL communications, namely $E_{k,h}^{(\mathrm{T})}=E_{k,0}^{(\mathrm{T})}+\gamma\cdot E_{0,h}^{(\mathrm{T})}$,
where $\gamma$ is the PUE of the BS or router hardware (if any).

\subsection{Tradeoff analysis in energy-constrained networks}

In what follows, we analyze the tradeoffs between meta-optimization
at the server and task-specific adaptation on the devices targeting
sustainable designs. As previously introduced, MAML requires high
energy costs $E_{\mathrm{ML}}$ (\ref{eq:eml}) for moving data over
UL. On the other hand, it simplifies subsequent task-specific model
adaptations, reducing the energy bill $E_{\mathrm{FL}}$ on the devices.
Communication and computing costs constitute the key indicators for
such optimal equilibrium: to simplify the analysis of (\ref{eq:eml})-(\ref{eq:cfa-1-1})
and (\ref{eq:efl})-(\ref{eq:fa-1}), energy costs are expressed $\forall k$
as efficiencies (bit/Joule) for uplink $\mathrm{E}_{\mathrm{UL}}=1/E_{k,0}^{(\mathrm{T})}$,
downlink $\mathrm{E}_{\mathrm{DL}}=1/E_{0,k}^{(\mathrm{T})}$, and
sidelink $\mathrm{E}_{\mathrm{SL}}=1/E_{k,h}^{(\mathrm{T})}$ communications
\cite{key-5}, as well as for computing (gradient per Joule or grad/J
for short) on the data center $\mathrm{E}_{\mathrm{0}}=1/E_{0}^{(\mathrm{C})}$
and for each \emph{k}-th devices $\mathrm{E}_{\mathrm{C}}=1/E_{k}^{(\mathrm{C})}$.
The problem we tackle is the minimization of the total energy cost
$\mathrm{E}$ required for the joint learning of $M$ tasks,
\begin{equation}
\mathrm{E}=E_{\mathrm{ML}}(t_{0},Q)+\sum_{i=1}^{M}E_{\mathrm{FL}}(t_{i}).\label{eq:budget}
\end{equation}
Besides communication and computing efficiencies, the energy budget
(\ref{eq:budget}) depends the required number of MAML rounds $t_{0}$
and the training tasks $Q$ chosen for meta-optimization, as well
as on the final accuracy, namely the number $t_{i}$ of FL rounds
implemented by the devices for task-specific refinements. 

\section{Application: deep reinforcement learning \label{sec:A-case-study}}

\begin{figure*}[!t]
\vspace{0.2cm}
\centering \includegraphics[scale=0.45]{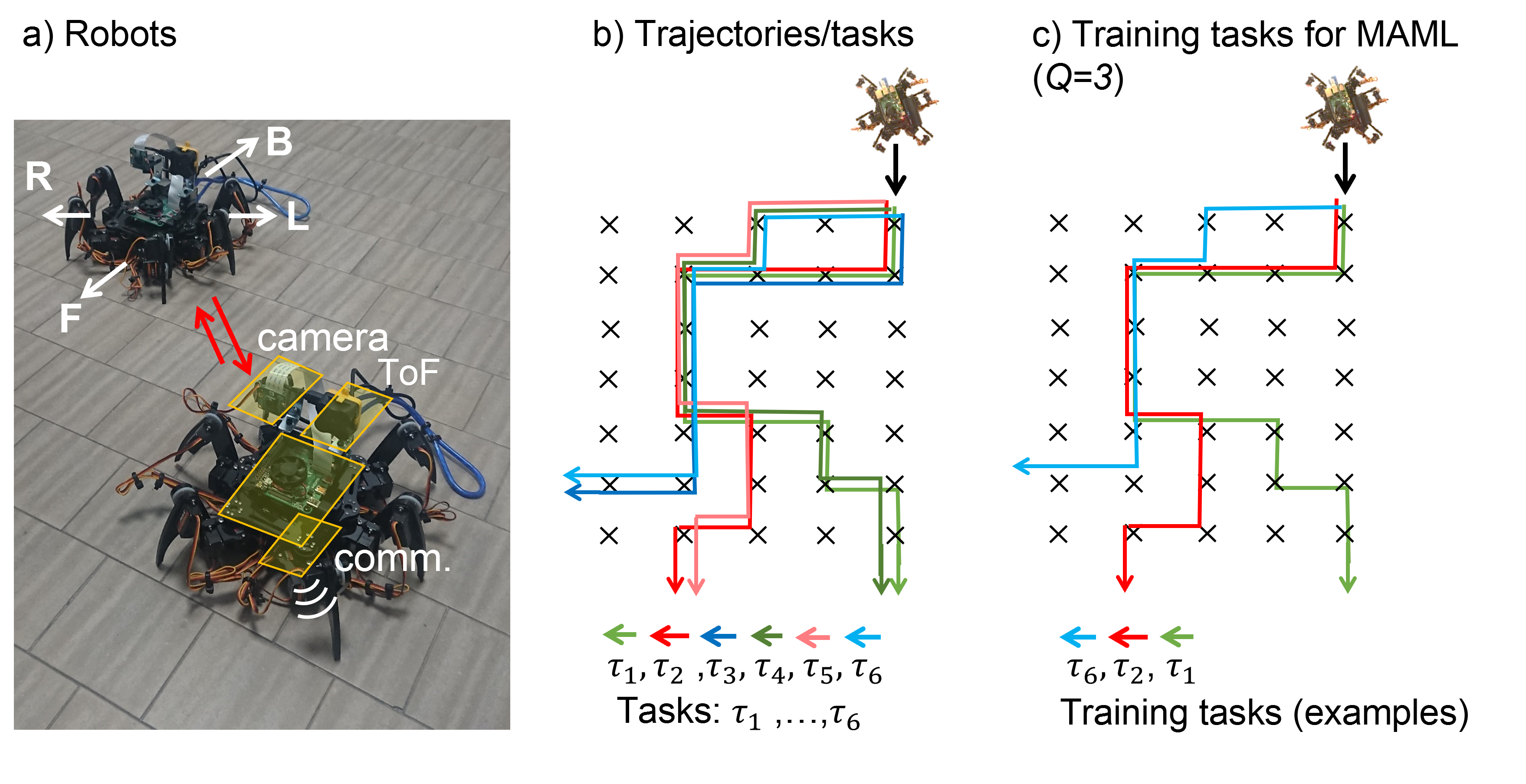} 
 \protect\caption{\label{comp1-1} MTL case study. From left to right: (a) crawling
robot deployment for data collection; (b) max-reward trajectories
corresponding to the $M=6$ tasks; (c) trajectories/tasks used for
MAML-based meta-model training.}
\vspace{-0.3cm}
\end{figure*}

The considered multitask DRL setting is depicted in Fig. \ref{comp1-1}(a).
Here, the agents are low-payload crawling robots organized into clusters:
the robots in each cluster can cooperate (via sidelink communications)
to learn a specific task. In particular, each cluster $\mathcal{C}_{i}$
is made up of 2 robots that collaborate to learn an optimized sequence
of motions (i.e., actions) to follow an assigned trajectory $\tau_{i}$,
namely, the task. A robot in cluster $\mathcal{C}_{i}$ that follows
the trajectory $\tau_{i}$ correctly has fulfilled the assigned task.
To simplify the setup, the trajectories followed by robots in each
cluster are chosen from $M=6$ pre-assigned ones, as shown in Fig.
\ref{comp1-1}(b). All trajectories have visible commonalities, i.e.
a common entry point, but with different exits (or paths to follow)
and are all implemented inside the same environment. 

The robots can independently explore the environment to collect training
data, namely state-action-reward sequences $\mathcal{E}_{i,k}$. However,
the motion control problem is simplified by forcing all robots to
move in a 2D regular grid space consisting of $40$ landmark points.
The action space ($y_{h,i}$) thus consists of $4$ motions: Forward
(F), Backward (B), Left (L), and Right (R). While moving in the grid
space, each robot collects state observations ($\mathbf{x}_{h,i}$)
obtained from two on-board cameras, namely a standard RGB camera and
a short-range Time Of Flight (TOF) one \cite{terabee}. Table \ref{parameters}
summarizes the relevant parameters for energy consumption evaluation.
The datasets used for the DRL process and the meta-learning system
are found in \cite{REPO}. Notice that the computing energy of the
data center and the devices, namely $P_{k}$ and $\mathrm{\mathrm{E}_{\mathrm{C}}}$,
are measured from the available hardware. On the other hand, we quantify
the estimated energy costs of meta-learning and FL stages by varying
the communication efficiencies $\mathrm{E}_{\mathrm{UL}},\mathrm{E}_{\mathrm{DL}},\mathrm{E}_{\mathrm{SL}}$.
Since real consumptions may depend on the specific protocol implementation
and be larger than the estimated ones, we will highlight the relative
comparisons.

\begin{table}[tp]
\protect\caption{\label{parameters} Main system parameters for energy footprint evaluation
on the data center (MAML) and on-devices (FL).}
\vspace{-0.2cm}

\begin{centering}
\begin{tabular}{lcc}
\toprule 
\textbf{Parameters} & \textbf{Data center ($k=0$)} & \textbf{Devices ($k\geq1$)}\tabularnewline
\midrule
\midrule 
Comp. $P_{k}$: & $590\,\mathrm{W}$$\,\,(350\,\mathrm{W}$ GPU) & $5.1\,\mathrm{W}$ (CPU)\tabularnewline
\midrule 
Batch \negthinspace{}time \negthinspace{}$T_{k}$: & $20\textrm{ ms}$ & $400\textrm{ ms}$\tabularnewline
\midrule 
Batches $B$: & $\mathrm{B}_{i}^{(a)},\mathrm{B}_{i}^{(b)}=10$ & $\mathrm{B}_{i}=20$\tabularnewline
\midrule 
Raw data size: & $Q\cdot b(\mathcal{E}_{i,k})$ MB & $b(\mathcal{E}_{i,k})\simeq24.6$ MB\tabularnewline
\midrule 
Model size: & $b(\mathbf{W})=5.6$ MB & $b(\mathbf{W})=5.6$ MB\tabularnewline
\midrule 
PUE $\gamma$: & $1.67$ & $1$\tabularnewline
\midrule 
Comp. $\mathrm{\mathrm{E}_{\mathrm{C}}}$: & $0.03$ grad/J & $0.16$ grad/J\tabularnewline
\bottomrule
\end{tabular}
\par\end{centering}
\medskip{}
 \vspace{-0.6cm}
\end{table}

\subsection{Multi-task learning and networking setup}

Each task $\tau_{i}$ is described by a position-reward lookup table
that assigns a reward value for each position in the 2D grid space,
according to the assigned trajectory. Maximum reward trajectories
for each of the $M=6$ tasks are detailed in Fig. \ref{comp1-1}(b).
Note that robots get a larger reward whenever they approach the desired
trajectory characterizing each task. 

The DeepMind model \cite{deepmind} is here used to represent the
parameterized Q-function $q_{\pi}(\mathbf{x},y|\mathbf{W})$ for DQL
implementation: it consists of $5$ trainable layers, and $1.3$ M
parameters with size $b(\mathbf{W})=5.6$ MB. The data center is remotely
located w.r.t. the area where the robots are deployed so that communication
is possible via cellular connectivity (UL/DL). The data center is
equipped with a CPU (Intel i7 8700K, $3.7$ GHz) and a GPU (Nvidia
Geforce RTX 3090, 24GB RAM). The robots mount a low-power ARM Cortex-A72
SoC and thus experience a larger batch time $T_{k}$, but a lower
power $P_{k}$ as reported in Table \ref{parameters}. In the following,
we assume that the robots can exchange the model parameters via SL
communication implemented by the WiFi IEEE 802.11ac protocol \cite{wifiaware}.

In each MAML round (described in Sect. \ref{sec:Energy-footprint-modeling}),
the data center collects observations $\mathcal{E}_{i,k}$ for $Q=3$
training tasks ($\tau_{1},\tau_{2},\tau_{6}$), as depicted in Fig.
\ref{comp1-1}(c). The training observations $\mathcal{E}_{i,k}:=(\mathbf{x}_{1,i},y_{1,i},r_{1,i},...,\mathbf{x}_{20,i},y_{20,i},r_{20,i})$
are obtained from $3$ robots and have size $b(\mathcal{E}_{i,k})\simeq24.6$
MB as corresponding to $20$ consecutive robot motions (using an $\varepsilon$-greedy
policy with $\varepsilon=0.1$). The training data are published to
the data center via UL with efficiency $\mathrm{E}_{\mathrm{UL}}$
by using the MQTT transport protocol. The MQTT broker is co-located
with the robots, while the data center retrieves the training sequences
by subscribing to the broker. Task-specific adaptation via FL requires
the robots in each cluster to mutually exchange the model parameters
on each round using SL with efficiency $\mathrm{E}_{\mathrm{SL}}$.
The MQTT payload is defined in \cite{access} and includes: \emph{i)}
the local model parameters $\mathbf{W}_{t,\tau_{i}}^{(k)}$ that are
binary encoded; \emph{ii)} the corresponding task $\tau_{i}$ description;
\emph{iii)} the FL round $t_{i}$ or learning epoch; \emph{iv)} the
average running reward $R$. For all tasks, the number of FL rounds
$t_{i}$ is selected to achieve the same average running reward of
$R=50$ that corresponds to learned trajectories with Root Mean Squared
Errors (RMSE) between $0.5$ m and $1$ m from the desired ones.

\begin{figure}[!t]
\centering \includegraphics[scale=0.65]{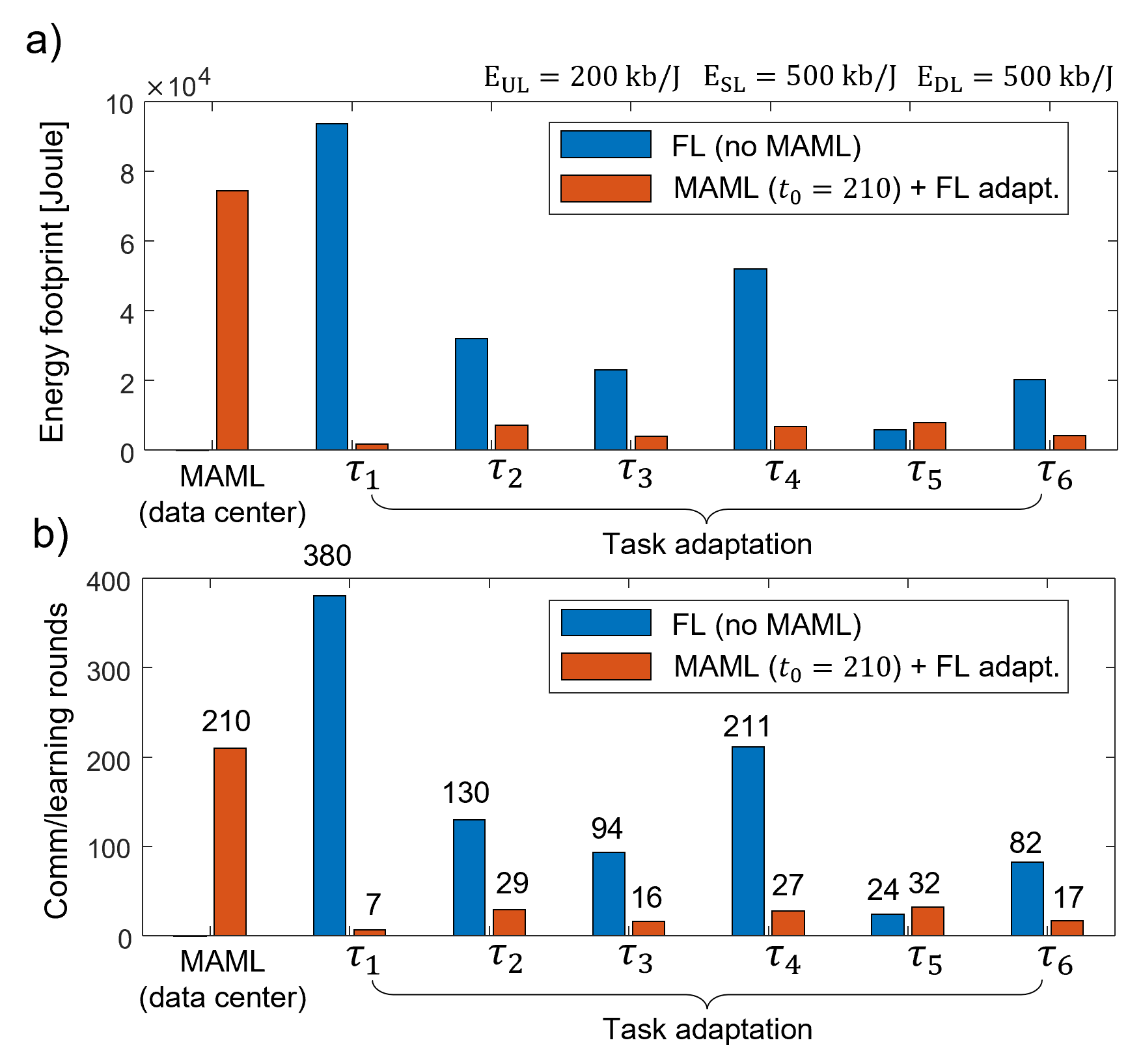} 
 \protect\caption{\label{maml+fl-1} Energy footprints and communication/learning rounds.
From top to bottom: (a) Energy footprint for MAML $E_{\mathrm{ML}}(t_{0})$,
and subsequent task adaptations $E_{\mathrm{FL}}(t_{i})$, $i=1,...,6$
(orange bars), compared with FL without MAML (blue bars). (b) Communication
and learning rounds required for MAML optimization ($t_{0}=210$ rounds
on the data center) and for specific tasks $t_{i}$, $i=1,...,6$
(orange bars), compared with FL without MAML (blue bars). Note that
the MAML energy cost per round in the data center (first bar from
left to right) is higher than the one on the devices.}
\vspace{-0.4cm}
\end{figure}

\subsection{FL vs. MAML: energy, communication and learning rounds}

Considering $M=6$ selected tasks, Fig. \ref{maml+fl-1}(a) shows
the energy costs required for MAML optimization on the data center,
and for the subsequent task-specific adaptations $\tau_{1},...,\tau_{6}$.
Fig. \ref{maml+fl-1}(b) depicts the corresponding number of rounds
($t_{i}$) required to achieve the running reward of $R=50.$ Other
learning parameters are defined in Table \ref{parameters}. In particular,
we consider a communication system characterized by $\mathrm{E}_{\mathrm{UL}}=200$
kb/J and $\mathrm{E}_{\mathrm{SL}}=500$ kb/J, in line with typical
WiFi IEEE 802.11ac implementations \cite{wifiaware,survey_energy_5G}.
All results are considered by averaging over $15$ different Monte
Carlo simulations using random model initializations (for both MAML
meta-optimization and subsequent FL). For the considered MAML scenario
(orange bars), the meta-model $\mathbf{W}_{t_{0},\tau}$ optimization
runs on the data center for $t_{0}=210$ rounds. $\mathbf{W}_{t_{0},\tau}$
is then transferred to the robots for task adaptation using decentralized
FL (\ref{eq:averaging}) over SL communications. In the same figure,
the energy footprints and the learning rounds are compared to those
obtained with a conventional approach (blue bars) where neither inductive
transfer nor MAML are implemented, while tasks are learned independently
by robots using decentralized FL \cite{key-5} with local models randomly
initialized (blue bars). 
\begin{figure}[!t]
	\centering \includegraphics[scale=0.60]{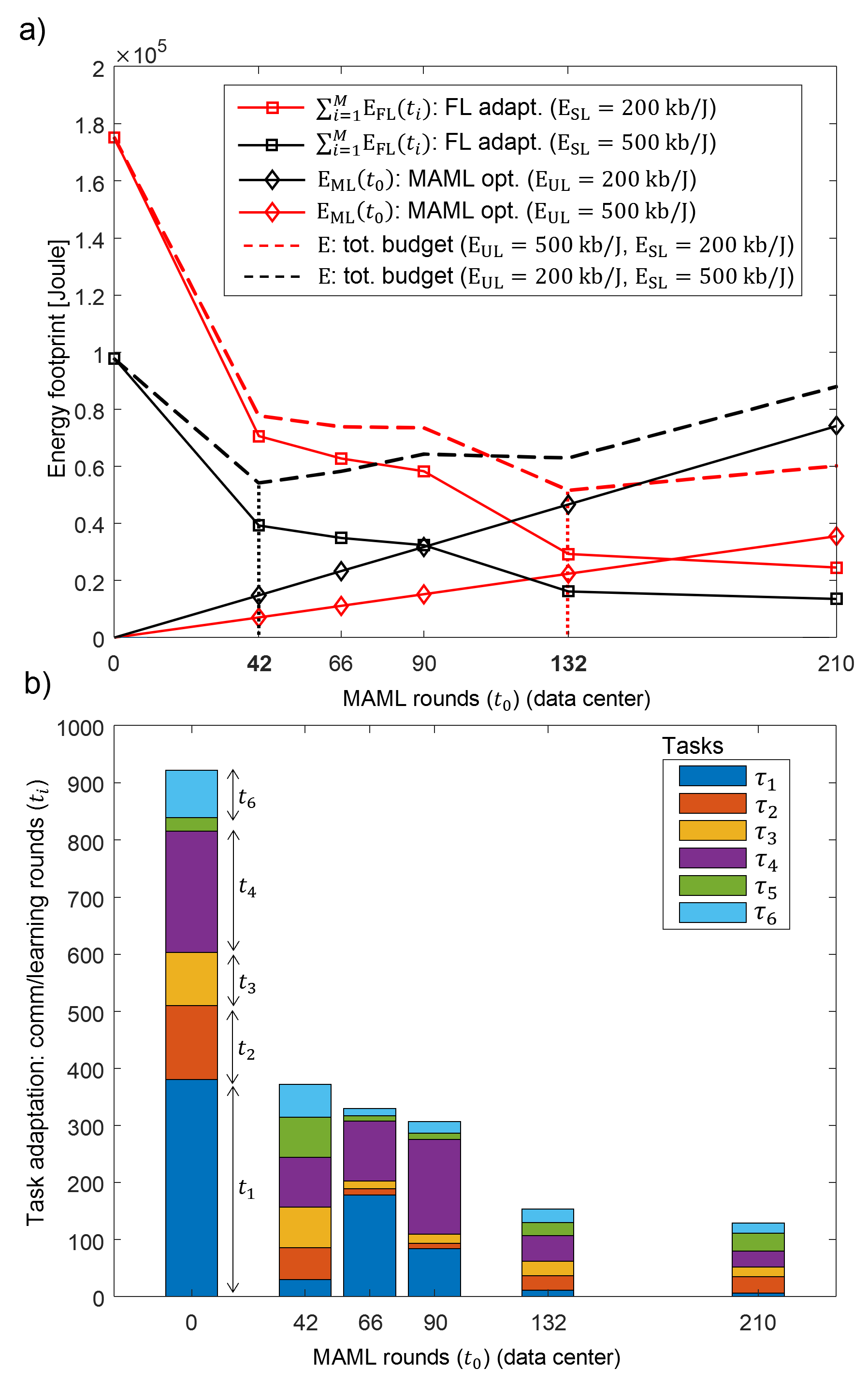} 
	\protect\caption{\label{maml+fl} Impact of MAML rounds $t_{0}$ on MTL for varying
		communication efficiencies. From top to bottom: (a) impact of MAML
		rounds $t_{0}$ on the meta-learning energy $E_{\mathrm{ML}}$ (diamond
		markers), task adaptations $\sum_{i=1}^{M}E_{\mathrm{FL}}(t_{i})$
		(squared markers), and total energy budget $\mathrm{E}$ in (\ref{eq:budget})
		in dashed lines. Black lines assumes D2D/mesh communications $\mathrm{E}_{\mathrm{SL}}=500$
		kb/J more efficient than UL, $\mathrm{E}_{\mathrm{UL}}=200$ kb/J,
		while red lines assume the opposite. (b) impact of $t_{0}$ on the
		number of FL rounds $t_{i}$ required for task $\tau_{1},...,\tau_{6}$
		adaptations.}
	\vspace{-0.1cm}
\end{figure}

MAML optimization is energy-hungry (with energy cost quantified as
$E_{\mathrm{ML}}=74$ kJ) as it requires a significant use of UL resources
for data collection over many learning rounds ($t_{0}=210$ in the
example) to produce the meta-model. On the other hand, as reported
in Fig. \ref{maml+fl-1}(b), task adaptations use few learning rounds/shots
$t_{i}$ for model refinements, namely ranging from $t_{1}=7$ to
$t_{5}=32$, reducing the robot energy footprints up to $10$ times
for all tasks, i.e., $E_{\mathrm{FL}}(t_{1})=1.6$ kJ, and $E_{\mathrm{FL}}(t_{5})=7.9$
kJ. Decentralized FL without inductive transfer unloads the data center
and requires marginal use of UL communication resources. However,
considering the same tasks, it requires much more FL rounds (from
$t_{5}=24$ to $t_{1}=380$) to converge compared with MAML approach.
Interestingly, using MAML inductive transfer for learning of task/trajectory
$\tau_{5}$ provides marginal benefits: this might be due to the fact
that learning of the specific task $\tau_{5}$ marginally benefits
from the knowledge of the meta-model. Overall, the total energy bill
required to learn all the $6$ tasks through MAML and FL for subsequent
task adaptation is quantified as $\mathrm{E}=E_{\mathrm{ML}}+\sum_{i=1}^{6}E_{\mathrm{FL}}(t_{i})=106$
kJ. This is approx. two times lower than the energy cost of learning
each task separately using only FL with no inductive transfer $\mathrm{E}=\sum_{i=1}^{6}E_{\mathrm{FL}}(t_{i})=227$
kJ.
\begin{table}[t]
	\vspace{0.5cm}
	\begin{tabular}{c|c|c|c|c|c|c|c|}
		\multicolumn{1}{c}{\multirow{2}{*}{}} &  & \multicolumn{6}{l|}{FL rounds $t_{i}$ per tasks $\tau_{i}$, $i=1,..,6$}\tabularnewline
		\cline{3-8} 
		&  & $t_{1}$ & $t_{2}$ & $t_{3}$ & $t_{4}$ & $t_{5}$ & $t_{6}$\tabularnewline
		\hline 
		\multirow{8}{*}{\begin{turn}{90}
				MAML rounds
		\end{turn}} & $t_{0}=0$  & \multirow{2}{*}{$380.1$} & \multirow{2}{*}{$129.6$ } & \multirow{2}{*}{$93.7$} & \multirow{2}{*}{$211.5$} & \multirow{2}{*}{$24.2$} & \multirow{2}{*}{$82.4$}\tabularnewline
		& (no MAML) &  &  &  &  &  & \tabularnewline
		\cline{2-8} 
		& $t_{0}=42$  & $29.7$ & $56.4$ & $70.9$ & $87$ & $70.4$ & $57.1$\tabularnewline
		\cline{2-8} 
		& $t_{0}=66$  & $178.8$ & $9.9$ & $14.3$ & $104.6$ & $9.8$ & $12.4$\tabularnewline
		\cline{2-8} 
		& $t_{0}=90$  & $84.9$ & $8.9$ & $15.6$ & $166.2$ & $11.3$ & $19.6$\tabularnewline
		\cline{2-8} 
		& $t_{0}=132$  & $11.6$ & $25.5$ & $25.1$ & $44.6$ & $23.1$ & $23.8$\tabularnewline
		\cline{2-8} 
		& $t_{0}=210$  & $6.7$ & $29.1$ & $16.5$ & $27.7$ & $32$ & $17.2$\tabularnewline
		\cline{2-8} 
		& $t_{0}=240$  & $2.7$ & $10.8$ & $9.1$ & $40$ & $21.8$ & $19.6$\tabularnewline
		\hline 
	\end{tabular}
	\medskip{}
	\protect\caption{\label{TAB} Average number of FL rounds $t_{i}$ for tasks $\tau_{1},...,\tau_{6}$,
		and varying $t_{0}$ as shown in Fig. \ref{maml+fl}(b). Average values
		w.r.t. 15 Monte Carlo runs.}
	\vspace{-0.6cm}
\end{table}
\subsection{MAML optimization and task adaptation tradeoffs}

Balancing MAML optimization on the data center, with task-specific
adaptations on the devices, is critical to improve efficiency. As
analyzed previously, MAML requires an initial high energy cost for
moving data on the UL over $t_{0}$ rounds; on the other hand, it
simplifies subsequent task-specific model adaptations, reducing the
energy bill for all tasks. Considering the same setting previously
analyzed, Fig. \ref{maml+fl} provides an in-depth analysis of MAML
and FL tradeoffs, for varying communication efficiencies. In particular,
in Fig. \ref{maml+fl}(a), we analyze the impact of MAML rounds $t_{0}$,
namely the split point between the meta-model and task-specific adaptation,
on the meta-learning energy cost $E_{\mathrm{ML}}$ (diamond markers),
on the subsequent task adaptations $\sum_{i=1}^{M}E_{\mathrm{FL}}(t_{i})$
(squared markers), and on the total energy budget $\mathrm{E}$ (\ref{eq:budget})
indicated in dashed lines. In Fig. \ref{maml+fl}(b) and Tab. \ref{TAB},
we quantify the corresponding number of rounds $t_{i}$ required for
task $\tau_{1},...,\tau_{6}$ adaptations and varying $t_{0}$. We
consider varying number of MAML rounds, namely $t_{0}:=\left\{ 42,66,90,132,210,240\right\} $,
each mapping onto a different number of SGD rounds for task-specific
training (\ref{nofederation}) and meta-model update (\ref{eq:met})
stages implemented on the server. 

In line with the results in Fig. \ref{maml+fl-1}, for the considered
settings, the use of meta-learning ($t_{0}>0$) generally cuts the
energy bill by least $2\times$ in all setups. Increasing the number
of rounds $t_{0}$ on the data center improves the meta-model generalization
capabilities and, as a result, reduces the energy cost (Fig. \ref{maml+fl}(a))
required for task adaptations $\sum_{i=1}^{M}E_{\mathrm{FL}}(t_{i})$
and the number of FL rounds $t_{i}$ (Fig. \ref{maml+fl}(b)). On
the other hand, moving data on the cloud for many rounds increases
the cost of MAML optimization $E_{\mathrm{ML}}$. 

A judicious system design taking into account both MAML and task adaptation
energy costs could bring significant energy savings. In Fig. \ref{maml+fl}(a)
we thus highlight the optimal number of rounds $t_{0}$ required to
minimize the total energy budget (\ref{eq:budget}). Optimal MAML
rounds depend on the uplink/sidelink communication costs: for example,
when $\mathrm{E}_{\mathrm{SL}}=500$ kb/J and $\mathrm{E}_{\mathrm{UL}}=200$
kb/J (black lines), the number of MAML rounds should be limited to
$t_{0}=42$, with $\mathrm{min_{\mathit{t_{\mathrm{0}}}}E}=56$ kJ.
On the opposite, more efficient UL than SL communications, namely
$\mathrm{E}_{\mathrm{UL}}=500$ kb/J and $\mathrm{E}_{\mathrm{SL}}=200$
kb/J (red lines), call for a larger number of MAML rounds ($t_{0}=132$)
to reduce the FL costs with $\mathrm{min_{\mathit{t_{\mathrm{0}}}}E}=52$
kJ. Besides energy costs, Table \ref{TAB} analyzes in more detail
the impact of $t_{0}$ on the number of FL rounds/shots $t_{i}$ required
for tasks $\tau_{1},...,\tau_{6}$ adaptations. The required rounds
for task learning without inductive transfer ($t_{0}=0$) sum to $\sum_{i=1}^{L}t_{i}=910$
(Fig. \ref{maml+fl}(b)) and scale down up to $9$ times ($\sum_{i=1}^{L}t_{i}=103$)
using $t_{0}$ MAML rounds on the data center. Notice that adaptations
to new tasks $\left\{ \tau_{3},\tau_{4},\tau_{5}\right\} \notin\mathcal{Q_{\tau}}$
not considered during meta-model training require (on average) more
rounds $\sum_{i=3,4,5}t_{i}=70.9$ compared with previously trained
tasks $\left\{ \tau_{1},\tau_{2},\tau_{6}\right\} \in\mathcal{Q_{\tau}}$,
$\sum_{i=1,2,6}t_{i}=33.1$. 

\section{Conclusions\label{sec:Conclusions}}

This work developed a novel framework for the energy footprint analysis
of Model-Agnostic Meta-Learning (MAML) geared towards Multi-Task Learning
(MTL) in wireless networks. The framework quantifies separately the
end-to-end energy costs when using a data center for the MAML optimization,
and the cost of subsequent task-specific model adaptations, implemented
using decentralized Federated Learning (FL). We examined novel trade-offs
about MAML optimization and few-shot learning over resource-constrained
devices. The analysis was validated in a multi-task RL setup where
robots collaborate to train an optimized sequence of motions to follow
different trajectories, or tasks. For the considered MTL setup, MAML
trains multiple tasks jointly and exploits task relationships to reduce
the energy bill by at least two times compared with FL without inductive
transfer. However, meta-learning requires moving data to the cloud
for many rounds as well as larger computing costs. In many cases,
the energy benefits of MAML also vary from task to task, suggesting
the need of optimized MAML stages taking into account the specific
commonalities among the tasks, including possible unseen ones. Finally,
depending on communication efficiencies, a judicious design of the
number of MAML rounds is critical to minimize the amount of data moved
to the cloud. Communication and learning co-design principles are
expected to further scale down the energy footprints.

\end{document}